\documentclass[review]{elsarticle}

\usepackage{hyperref}
\usepackage{color}
\usepackage{amsmath,amssymb,amsfonts,amsthm,enumerate,multirow,mathtools}
\usepackage{color}
\usepackage{siunitx}
\usepackage{graphicx}
\usepackage{parskip}
\usepackage{placeins} 
\usepackage{booktabs}
\usepackage{algorithm,algpseudocode}
\usepackage[numbers]{natbib}
\usepackage{subcaption}
\usepackage{cancel}
\definecolor{darkred}{RGB}{139,0,0}

\usepackage{tikz}
\usetikzlibrary{positioning}
\tikzset{%
  every neuron/.style={
    circle,
    draw,
    minimum size=1cm
  },
  neuron missing/.style={
    draw=none, 
    scale=4,
    text height=0.333cm,
    execute at begin node=\color{black}$\vdots$
  },
  every neuron2/.style={
    rectangle,
    draw,
    minimum size=1cm
  }
}

\newcommand{\pmat}[1]{\begin{bmatrix}#1\end{bmatrix}}

\newcommand{\Transp}{\mathsf{T}}
\newcommand{\red}[1]{{\color{black}#1}}

\newcommand{\pd}[2]{\frac{\partial #1}{\partial #2}}
\newcommand{\pdd}[2]{\frac{\partial^2 #1}{\partial #2^2}}
\newcommand{\pdt}[3]{\frac{\partial^2 #1}{\partial #2 \partial #3 }}

\usepackage[scr=boondoxo,scrscaled=1.05]{mathalfa}
\def\Fcal{\mathcal{C}} %
\def\Fcalmat{\mathcal{C}} %
\def\Gcalmat{\mathcal{G}} %

\renewcommand{\vec}[1]{%
    \mathbf{\boldsymbol{\mathrm{#1}}}
}
\def\x{\vec{x}}

\journal{Journal of Computational Physics}

\begin{document}

\begin{frontmatter}

\title{Linearly Constrained Neural Networks}


\author[uon]{Johannes~N.~Hendriks\corref{mycorrespondingauthor}}
\cortext[mycorrespondingauthor]{Corresponding author}
\ead{johannes.hendriks@newcastle.edu.au}

\author[UU]{Carl~Jidling}
\author[uon]{Adrian~G.~Wills}
\author[UU]{Thomas~B.~Sch\"{o}n}

\address[uon]{School of Engineering, The University of Newcastle, Callaghan NSW 2308, Australia}
\address[UU]{Department of Information Technology, Uppsala University, 75105 Uppsala, Sweden}

\begin{abstract}
\red{We present a novel approach to modelling and learning vector fields from physical systems using neural networks that explicitly satisfy known linear operator constraints.}
To achieve this, the target function is modelled as a linear transformation of an underlying potential field, \red{which is in turn modelled by a neural network.}
This transformation is chosen such that any prediction of the target function is guaranteed to satisfy the constraints. 
The approach is demonstrated on both simulated and real data examples.
\end{abstract}


\end{frontmatter}


\section{Introduction} 
\label{sec:introduction}
Developments during recent years have established deep learning as the perhaps most prominent member of the machine learning toolbox.
Today neural networks are present in a broad range of applications and are used for both classification and regression problems.  
\red{This includes the use of neural networks to model and learn vector-valued quantities from physical systems such as magnetic fields \cite{pukrittayakamee2009simultaneous}, plasma fields \cite{van1995neural}, and the dynamics of conservative systems \cite{greydanus2019hamiltonian,Lutter2019DeepLN}, to name a few.}
The popularity of neural networks is to a large extent explained by the highly flexible nature that enables these models to encode a very large class of non-linear functions.

Nevertheless, the performance of the neural network is often dependent on careful design and the amount of training data available.
In particular, a larger network is more flexible but also requires more training data to reduce the risk of overfitting.   
Different types of regularisation techniques are sometimes used to facilitate this balance.

\begin{figure}[!ht]
    \centering
     \includegraphics[width=0.7\linewidth]{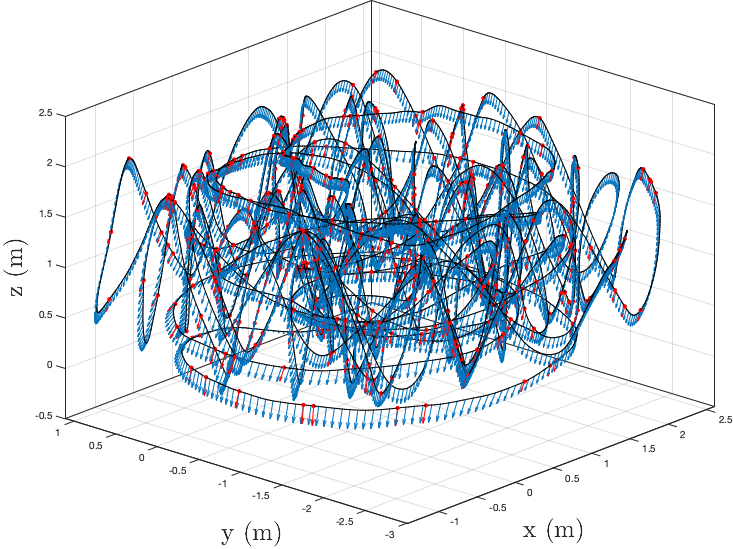}
    \caption{Magnetic field predictions (blue) using a constrained neural network trained on 500 observations (red) sampled from the trajectory indicated by the black curve. The magnetic field, $\mathbf{B}$ is curl-free satisfying the constraint $\nabla \mathbf{B}$ = 0, and the method proposed in this paper ensures that the predictions satisfy this constraint.}
    \label{fig:mag_predictions}
\end{figure}

Instead of focusing on the network \textit{per se}, it may be just as important to consider prior knowledge provided by the problem setting.
For instance, the function of interest can represent a quantity subject to fundamental physical constraints.
In some cases these physical constraints take the form of linear operator constraints.
This includes many vector fields that are known to be either divergence- or curl-free.
Examples of divergent-free vector fields (also known as solenoidal fields) are the magnetic field \citep{konopinski1978electromagnetic}---see Figure~\ref{fig:mag_predictions}, the velocity of an anelastic flow \citep{durran1989improving}, the vorticity field 
\citep{kundu2015fluid,truesdell2018kinematics}, and current density where the charge is constant over time as given by the continuity equation \citep{chow2006introduction,griffiths1962introduction}.
Low-mach number flow is a simplification of the compressible Euler equations and describes flow with a prescribed divergence \citep{almgren2006low} (i.e. it takes the form of an affine constraint).
Another example of fields satisfying an affine constraint is given by Maxwell's equations, which describe electromagnetic fields with a prescribed curl and divergence \citep{fleisch2008student}.
Within continuum mechanics, the stress field and strain field inside a solid object satisfy the equilibrium conditions and the strain field inside a simply connected body satisfies the compatibility constraints \citep{sadd2009elasticity}.
\red{Equilibrium can also be used as a constraint when modelling plasma fields \cite{van1995neural}.}

The list can be made longer, but the point is clear -- by making sure that certain constraints are fulfilled, we (significantly) reduce the set of functions that \textit{could} explain our measured data.
This, in turn, implies that we can maintain high performance without requiring the same amount of flexibility. 
Put simply: we can obtain the same results with a smaller network and less training data.

\subsection{Contribution} 
\label{sec:contribution}


This paper presents a novel approach for designing neural network based models that satisfy linear operator constraints.
\red{It is a step towards addressing a central issue applying machine learning methods to physical systems: enforcement of physical constraints in machine learning techniques.
The approach models the vector field to be learned as a linear transformation of an underlying potential field.}
The benefits of using this approach are two-fold:
\begin{enumerate}
  \item Any predictions made using this approach will satisfy the constraints for the entire input space.
  \item Incorporating known constraints reduces the problem size. This reduces the amount of training data required and also allows a smaller neural network to be used while still achieving the same performance.
  Reducing the amount of data required can save time and money during the data collection phase.
\end{enumerate}
\red{Additionally, the approach allows a wide class of neural network models to be used to model the underlying function, including convolutional neural networks and recurrent neural networks.}

Existing methods (see Section~\ref{sec:related_work}---Related Work), have predominately tackled the problem by either (a) augmenting the cost function to penalise constraint violation or (b) developing problem specific models. In contrast we present a general approach that guarantees the constraints to be satisfied, and can be used for any linear operator constraints. \red{Further, we show that the proposed approach has better performance than that given by augmenting the cost function --- a standard approach to incorporating constraints into neural network models.}

\section{Problem Formulation} 
\label{sec:problem_formulation}
Assume we are given a data set of $N$ observations $\{\mathbf{x}_i,\mathbf{y}_i\}_{i=1}^N$ where $\mathbf{x}_i$ denotes the input and $\mathbf{y}_i$ denotes the output. Both the input and output are potentially vector-valued with $\mathbf{x}_i\in\mathbb{R}^D$ and $\mathbf{y}_i\in\mathbb{R}^K$. 
Here we consider the regression problem where the data can be described by the non-linear function $\mathbf{y}_i = \mathbf{f}(\mathbf{x}_i) + \mathbf{e}_i$, where $\mathbf{e}_i$ is zero-mean white noise representing the measurement uncertainty. 
In this work, a neural network is used to model $\mathbf{f}$ and can be described by
\begin{equation}
    \mathbf{f}(\mathbf{x}) = \mathbf{h}_L(\mathbf{h}_{L-1}(\cdots \mathbf{h}_2(\mathbf{h}_1(\mathbf{x})))),
\end{equation}
where each $\mathbf{h_l}(\mathbf{z})$ has the form
\begin{equation}
  \mathbf{h}_l(\mathbf{z}) = \phi_l(\mathbf{W}_l\mathbf{z}+\mathbf{b}_l).
\end{equation}
Here, $L$ is the number of layers in the neural network, each $\phi_l$ is an element-wise non-linear function commonly referred to as an activation function and $\{\mathbf{W}^l,\mathbf{b}^l\}_{l=1}^{L}$ are the parameters of the neural network that are to be learned from the data.

In addition to the data, we know that the function $\mathbf{f}$ should fulfil certain constraints
\begin{equation}\label{eq:constraints}
    \mathcal{C}_\mathbf{x}[\mathbf{f}] = \mathbf{0},
\end{equation}
where $\mathcal{C}_\mathbf{x}$ is a linear operator \cite{luenberger1997optimization} mapping the function $\mathbf{f}$ to another function $\mathbf{g}$. That is $\mathcal{C}_\mathbf{x}[\mathbf{f}] = \mathbf{g}$. Further, we restrict $\mathcal{C}_\mathbf{x}$ to be a linear operator, meaning that $\mathcal{C}_\mathbf{x}[\lambda_1\mathbf{f}_1+\lambda_2\mathbf{f}_2] = \lambda_1\mathcal{C}_\mathbf{x}[\mathbf{f}_1]+\lambda_2\mathcal{C}_\mathbf{x}[\mathbf{f}_2]$, where $\lambda_1,\lambda_2\in\mathbb{R}$. A simple example is if the operator is a linear transformation $\mathcal{C}_\mathbf{x}[\mathbf{f}] = \mathbf{C}\mathbf{f}$ which together with the constraints \eqref{eq:constraints} forces a certain linear combination of the outputs to be linearly dependent.

\red{The operator $\mathcal{C}_\mathbf{x}$ can be used to represent a wide class of linear constraints on the function $\mathbf{f}$ and for a background on linear operators the interested reader is referred to \cite{luenberger1997optimization}.}
For example, we might know that the function $\mathbf{f}:\mathbb{R}^2\to\mathbb{R}^2$ should obey the partial differential equation $\mathcal{C}_\mathbf{x}[f] = \pd{f_1}{x_1} + \pd{f_2}{x_2} = 0$.

The constraints can come from either known physical laws or other prior knowledge about the data. Here, the objective is to determine an approach to derive models based on neural networks such that all predictions from these models will satisfy the constraints.

\section{Building a Constrained Neural Network} 
\label{sec:building_a_constrained_neural_network}
In this section, an approach to learn a function using a neural network such that any resulting estimate satisfies the constraints \eqref{eq:constraints} is proposed.
\red{The novelty of this approach is in recognising that vector fields subject to linear constraints can, in general, be modelled by a linear transformation of a neural network such that the learned field will always satisfy the constraints.
This section first presents the approach and then gives a brief discussion of conditions that may be imposed on the neural network.
Then, the following section gives two methods for determining the required transformation.}

\subsection{Our Approach} 
\label{sub:our_approach}
Our approach designs a neural network that satisfies the constraint for all possible values of its parameters, rather than imposing constraints on the parameter values themselves.
This is done by considering $\mathbf{f}$ to be related to another function $\mathbf{g}$ via some linear operator $\mathcal{G}_\mathbf{x}$:
\begin{equation}\label{eq:potential_mapping}
    \mathbf{f} = \mathcal{G}_\mathbf{x}[\mathbf{g}].
\end{equation}
The constraints \eqref{eq:constraints} can then be written as 
\begin{equation}\label{eq:potential_constraints}
    \mathcal{C}_{\mathbf{x}}[\mathcal{G}_\mathbf{x}[\mathbf{g}]] = \mathbf{0}.
\end{equation}
We require this relation to hold for any function $\mathbf{g}$. To do this, we will interpret $\mathcal{C}_\mathbf{x}$ and $\mathcal{G}_\mathbf{x}$ as matrices and use a similar procedure to that of solving systems of linear equations. Since $\mathcal{C}_\mathbf{x}$ and $\mathcal{G}_\mathbf{x}$ are linear operators, we can think of $\mathcal{C}_\mathbf{x}[\mathbf{f}]$ and $\mathcal{G}_\mathbf{x}[\mathbf{g}]$ as matrix-vector multiplications where $\mathcal{C}_\mathbf{x}[\mathbf{f}] = \mathcal{C}_\mathbf{x}\mathbf{f}$, with $(\mathcal{C}_\mathbf{x}\mathbf{f})_i = \sum_{j=1}^K (\mathcal{C}_\mathbf{x})_{ij}\mathbf{f}_j$ where each element $(\mathcal{C}_\mathbf{x})_{ij}$ in the operator matrix is a scalar operator. With this notation, \red{\eqref{eq:potential_constraints} can be expressed as matrix-vector products \cite{luenberger1997optimization}:}
\begin{equation}
    \mathcal{C}_\mathbf{x}\mathcal{G}_\mathbf{x}\mathbf{g} = \mathbf{0},
\end{equation}
where a solution is given by 
\begin{equation}\label{eq:operator_sol}
    \mathcal{C}_\mathbf{x}\mathcal{G}_\mathbf{x} = \mathbf{0}.
\end{equation}
This reformulation imposes constraints on the operator $\mathcal{G}_\mathbf{x}$ rather than on the neural network model of $\mathbf{f}$ directly. We can then proceed by first modelling the function $\mathbf{g}$ as a neural network and then transform it using the mapping \eqref{eq:potential_mapping} to provide a neural network for $\mathbf{f}$ that explicitly satisfies the constraints according to
\begin{equation}\label{eq:operator_NN}
    \mathbf{f} = \mathcal{G}_{\mathbf{x}}\mathbf{g}.
\end{equation}

\begin{figure}[!ht]
    \centering
    \subcaptionbox{Standard Neural Network \label{fig:SNN}}{
    \begin{tikzpicture}[x=1.25cm, y=1.25cm, >=stealth,scale=0.7, every node/.style={transform shape}]
\foreach \m/\l [count=\y] in {1,2,3,missing,4}
  \node [every neuron/.try, neuron \m/.try] (input-\m) at (0,2.5-\y) {};

\foreach \m [count=\y] in {1,missing,2}
  \node [every neuron/.try, neuron \m/.try ] (hidden-\m) at (1.5,2-\y*1.25) {};

\foreach \m [count=\y] in {1,2,missing,n}
  \node [every neuron/.try, neuron \m/.try ] (output-\m) at (3,1.5-\y) {};

\foreach \l [count=\i] in {1,2,3,n}
  \node at (input-\i.center) {$x_\l$};

\foreach \l [count=\i] in {1,n}
  \node at (hidden-\i.center) {$a_\l$};

\foreach \l [count=\i] in {1,2,n}
  \node at (output-\l.center) {$f_\l$};

\foreach \i in {1,...,4}
  \foreach \j in {1,...,2}
    \draw [->] (input-\i) -- (hidden-\j);

\foreach \i in {1,...,2}
  \foreach \j in {1,2,n}
    \draw [->] (hidden-\i) -- (output-\j);

\foreach \l [count=\x from 0] in {Input, Hidden, Ouput}
  \node [align=center, above] at (\x*1.5,2) {\l \\ layer};
\end{tikzpicture}}
\subcaptionbox{Our Constrained Neural Network \label{fig:CNN}}{
\begin{tikzpicture}[x=1.25cm, y=1.25cm, >=stealth,scale=0.7, every node/.style={transform shape}]
\foreach \m/\l [count=\y] in {1,2,3,missing,4}
  \node [every neuron/.try, neuron \m/.try] (input-\m) at (0,2.5-\y) {};

\foreach \m [count=\y] in {1,missing,2}
  \node [every neuron/.try, neuron \m/.try ] (hidden-\m) at (1.5,2-\y*1.25) {};

\foreach \m [count=\y] in {1}
  \node [every neuron/.try, neuron \m/.try ] (output-\m) at (3,1.5-\y) {};

\foreach \m [count=\y] in {1}
  \node [every neuron2/.try, neuron \m/.try ] (mapping-\m) at (4.5,1.5-\y) {$\mathcal{G}_\mathbf{x}$};

\foreach \m [count=\y] in {1,2,missing,n}
  \node [every neuron/.try, neuron \m/.try ] (func-\m) at (6,1.5-\y) {};

\foreach \l [count=\i] in {1,2,3,n}
  \node at (input-\i.center) {$x_\l$};

\foreach \l [count=\i] in {1,n}
  \node at (hidden-\i.center) {$a_\l$};

\foreach \l [count=\i] in {1}
  \node at (output-\i.center) {$g$};

\foreach \l [count=\i] in {1,2,n}
  \node at (func-\l.center) {$f_\l$};

\foreach \i in {1,...,4}
  \foreach \j in {1,...,2}
    \draw [->] (input-\i) -- (hidden-\j);

\foreach \i in {1,...,2}
  \foreach \j in {1}
    \draw [->] (hidden-\i) -- (output-\j);

\foreach \i in {1}
  \foreach \j in {1}
    \draw [->] (output-\i) -- (mapping-\j);

\foreach \i in {1}
  \foreach \j in {1,2,n}
    \draw [->] (mapping-\i) -- (func-\j);

\foreach \l [count=\x from 0] in {Input, Hidden, Ouput}
  \node [align=center, above] at (\x*1.5,2) {\l \\ layer};
\node [align=center, above] at (4.5,2) {Mapping \\ function};
\node [align=center, above] at (6,2) {Constrained \\ Outputs}; 
\end{tikzpicture}}
    \caption{Diagram illustrating the difference between a standard neural network structure (\subref{fig:SNN}) and our constrained model (\subref{fig:CNN}). Since the output layer of the constrained model is of lower dimension than that of the standard neural network, less hidden layers and neurons are required. In this figure, we assume that $g$ is scalar and show a single hidden layer.}
    \label{fig:latent_ill}
\end{figure}
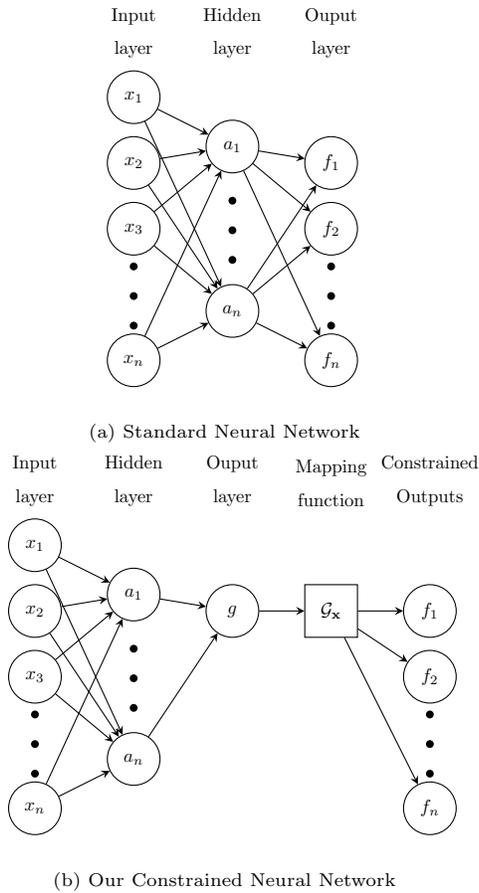

An illustration of the constrained model is given in Figure~\ref{fig:latent_ill}. 
The procedure to design the neural network can now be divided into three steps:
\begin{enumerate}
    \item Find an operator $\mathcal{G}_\mathbf{x}$ satisfying the condition \eqref{eq:operator_sol}.
    \item Choose a neural network structure for $\mathbf{g}$.
    \item Determine the neural network based model for $\mathbf{f}$ according to \eqref{eq:operator_NN}.
\end{enumerate}

The choice of neural network structure in step 2 may have some conditions placed upon it by the transformation found in step 1 for the resulting model to be mathematically correct. For example, if the transformation contains partial derivatives then this may restrict the choice of activation function. A more detailed discussion is given in Section~\ref{sec:conditions_on_the_neural_network_to_satisfy_derivative_constraints}.
\red{Despite these conditions, the approach admits a reasonably wide class of neural network models including fully connected neural networks (FCNN), convolutional neural networks (CNN), and recurrent neural networks (RNN). For the ease of explanation, the examples given in this paper use FCNNs.}

The parameters of the resulting model can be learned using existing methods such as stochastic gradient descent.
It is worth noting that if the data requires scaling then care should be taken as this scaling can modify the form of the constraints.

In the case where the operator $\mathcal{G}_\mathbf{x}$ contains partial derivatives, such as for curl-free and divergent-free fields, the implementation can be done using automatic differentiation such as the \texttt{grad} function in \texttt{PyTorch} \citep{paszke2017automatic}. 



\subsection{Conditions due to Derivative Transformations} 
\label{sec:conditions_on_the_neural_network_to_satisfy_derivative_constraints}
When the transformation $\mathcal{G}_\mathbf{x}$ contains partial derivatives the underlying neural network $\mathbf{g}$ must be chosen to satisfy some conditions.
Intuitively, it is required that the partial derivative of the neural network must be a function of both the inputs and the network parameters. If this is not the case, then the model loses the ability to represent a spatially varying target function. 
Here, we provide a few examples of this. \\

If the transformation contains only first-order derivatives then this does not result in any restrictive conditions. To see this consider a neural network with a single hidden layer and an identity activation function in its output layer, written along with its partial derivative as
\begin{equation}
\begin{split}
    \mathbf{g}(\mathbf{x}) &= \mathbf{W}_2\phi_1(\mathbf{W}_1\mathbf{x}+\mathbf{b}_1)+\mathbf{b}_2 = \mathbf{W}_2\phi_1(\mathbf{a}_1)+\mathbf{b}_2,\\\
    \pd{\mathbf{g}(\mathbf{x})}{\mathbf{x}} &= \mathbf{W}_2\pd{\mathbf{a}_1}{\mathbf{x}}\pd{\phi_1(\mathbf{a}_1)}{\mathbf{a}_1} = \mathbf{W}_2\mathbf{W}_1\pd{\phi_1(\mathbf{a}_1)}{\mathbf{a}_1}.
\end{split}
\end{equation}
Here, we have introduced the notation $\mathbf{a}_1 = \mathbf{W}_1\mathbf{x}+\mathbf{b}_1$ to simplify the description.
Hence, it is only required that the first derivative of the activation function with respect to $\mathbf{a}_1$ is not constant. However, for higher-order derivatives, there are requirements on the activation functions chosen. 
Consider the second derivatives of the same network
\begin{equation}
\begin{split}
    \pdd{\mathbf{g}(\mathbf{x})}{\mathbf{x}} &= \mathbf{W}_2\left(\cancelto{0}{\pdd{\mathbf{a}_1}{\mathbf{x}}}\pd{\phi_1(\mathbf{a}_1)}{\mathbf{a}_1}+ \pd{\mathbf{a}_1}{\mathbf{x}}\pdd{\phi_1(\mathbf{a}_1)}{\mathbf{a}_1}\right) 
    = \mathbf{W}_2\pd{\mathbf{a}_1}{\mathbf{x}}\pdd{\phi_1(\mathbf{a}_1)}{\mathbf{a}_1}.
\end{split}
\end{equation}
To use this model it is required that the second derivatives of the activation function are non-constant. This excludes, for instance, the \texttt{ReLU} function. 
The same procedure can be easily used to show that this condition remains when the neural network is extended to two hidden layers.

\section{Finding the Transformation Operator} 
\label{sec:finding_the_projection_operator}
This section presents two methods for determining a suitable operator $\mathcal{G}_\mathbf{x}$.
Prior knowledge about the physics of a problem could inform the choice of operator. If this is not the case, then a suitable operator could be found by proposing an ansatz and solving a system of linear equations.

\subsection{From Physics} 
\label{sub:from_physics}
From fundamental physics, it may be the case that we know that the vector field of interest is related to an underlying potential field. 
Common examples of this are divergence-free ($\nabla\cdot\mathbf{f} = 0$)\footnote{$\nabla = \pmat{\pd{}{x} &\pd{}{y} & \pd{}{z}}^{\Transp}$.} vector fields and curl-free ($\nabla\times\mathbf{f} = 0$) vector fields.
A curl-free vector field can be written as a function of an underlying scalar potential field $g$:
\begin{equation}
    \mathbf{f} = \nabla\cdot g,
\end{equation}
which gives $\mathcal{G}_\mathbf{x} = \pmat{\pd{}{x} &\pd{}{y} & \pd{}{z}}^{\Transp}$.
Divergence-free vector fields can, on the other hand, be expressed as a function of a vector potential field $\mathbf{g}\in\mathbb{R}^3$, given by
\begin{equation}
    \mathbf{f} = \nabla\times\mathbf{g},
\end{equation}
which gives
\begin{equation}
    \mathcal{G}_\mathbf{x} = \pmat{0 & -\pd{}{z} & \pd{}{y}\\
                                \pd{}{z} & 0 & -\pd{}{x} \\
                                -\pd{}{y} & \pd{}{x} & 0}.
\end{equation}

Many natural phenomena can be modelled according to these constraints, and several examples were given in Section~\ref{sec:introduction}.


\subsection{Ansatz} 
\label{sub:anzats}
In absence or ignorance of underlying mathematical relations, the operator $\Gcalmat_\x$ can be constructed using the pragmatic approach of which an exhaustive version is described by Jidling et. al.~\citep{jidling2017linearly}; a brief outline is given below.
A solid analysis of the mathematical properties of this operator is provided by Lange-Hegermann~\citep{lange-hegermann2018algorithmic}.

The cornerstone of the approach is an ansatz on what operators we assume $\Gcalmat_\x$ to contain; we formulate it as
\begin{align}\label{eq:anzats}
\Gcalmat_\x = \Gamma\vec{\xi},
\end{align} 
where $\vec{\xi}$ is a vector of operators, and $\Gamma=[\gamma_{ij}]$ is a real-valued matrix that we wish to determine.
Here, we have assumed for simplicity that $\Gcalmat_\x$ is a vector, implying that $g$ is a scalar function.
We now use~\eqref{eq:anzats} to rewrite \eqref{eq:operator_sol} as
\begin{align}
\Fcal_\x\Gamma\vec{\xi} = \vec{0}.
\end{align}   
  Expanding the product on the left-hand side, we find that it reduces to a linear combination of operators.
  Requiring all coefficients to equal 0, we obtain a system of equations from which we can determine $\Gamma$, and thus also $\Gcalmat_\x$.

  For illustration, consider a toy example where
  \begin{align}
  \Fcalmat_\x = 
  \begin{bmatrix}
  \pd{}{x} & \pd{}{y}
  \end{bmatrix}.
  \end{align}
  Assuming that $\Gcalmat_\x$ contains the same operators as $\Fcalmat_\x$, we let 
  \begin{align}
  \vec{\xi} = 
  \begin{bmatrix}
  \pd{}{x} & \pd{}{y}
  \end{bmatrix}^\Transp.
  \end{align}
  We then expand

  Requiring this expression to equal 0, we get the following system of equations
  \begin{equation}
  \begin{bmatrix}
  1 & 0 & 0 & 0 \\
  0 & 1 & 1 & 0 \\
  0 & 0 & 0 & 1
  \end{bmatrix}
  \begin{bmatrix}
  \gamma_{11} \\
  \gamma_{12} \\
  \gamma_{21} \\
  \gamma_{22}
  \end{bmatrix}
  =\vec{0},
  \end{equation}
  which is solved by $\gamma_{11}=\gamma_{22}=0$ and $\gamma_{12}=-\gamma_{21}$.
  Letting $\gamma_{21}=1$, we obtain
  \begin{align}
  \Gcalmat_\x=
  \begin{bmatrix}
  -\pd{}{y} & \pd{}{x}
  \end{bmatrix}^\Transp,
  \end{align}
  which can easily be verified to satisfy~\eqref{eq:operator_sol}.
  
  In the general case, $\Gcalmat_\x$ may contain operators of higher order than those in $\Fcalmat_\x$.
  It is also possible that a suitable underlying function may have a vector rather than a scalar output.
  The procedure should, therefore, be considered iterative. 
  Additionally, within \ref{sec:simulated_affine_example}, we show that this approach can be extended to affine constraints.
  

\section{Experimental Results} 
\label{sec:experimental_results}
In this section, we demonstrate the proposed approach on simulated data from a divergence-free field, simulated data of a strain field satisfying the equilibrium conditions, and real data of a magnetic field that satisfies the curl-free constraint. Python code to run the examples in this section is available at \url{https://github.com/jnh277/Linearly-Constrained-NN/}.

\red{These demonstrations use an FCNN for the underlying model in the proposed approach. Hence, they compare the results to a standard FCNN. Additionally, we compare the performance of the proposed approach and the commonly used approach of augmenting the cost function to penalise constraint violations.}

\subsection{Simulated Divergence-Free Function} 
\label{sub:simulated_divergence_free_function}
Consider the problem of modelling a divergence-free vector field defined as
\small
\begin{equation}
\begin{split}
    f_1(x_1,x_2) &= \exp(-ax_1x_2)(ax_1\sin(x_1x_2)-x_1\cos(x_1x_2)), \\
    f_2(x_1,x_2) &= \exp(-ax_1x_2)(x_2\cos(x_1x_2)-ax_2\sin(x_1x_2)),
\end{split}
\end{equation}
\normalsize
where $a$ is a constant. This vector field satisfies the constraint $\pd{f_1}{x_1}+\pd{f_2}{x_2} = 0$.
A neural network based model satisfying these constraints is given by
\begin{equation}
    \mathbf{f} = \pmat{\pd{}{x_2} \\ -\pd{}{x_1}}g.
\end{equation}

\begin{figure*}[tb]
    \centering
    \subcaptionbox{Data size study\label{fig:n_data_study}}{\includegraphics[trim=90 5 150 5, clip,width=0.45\linewidth]{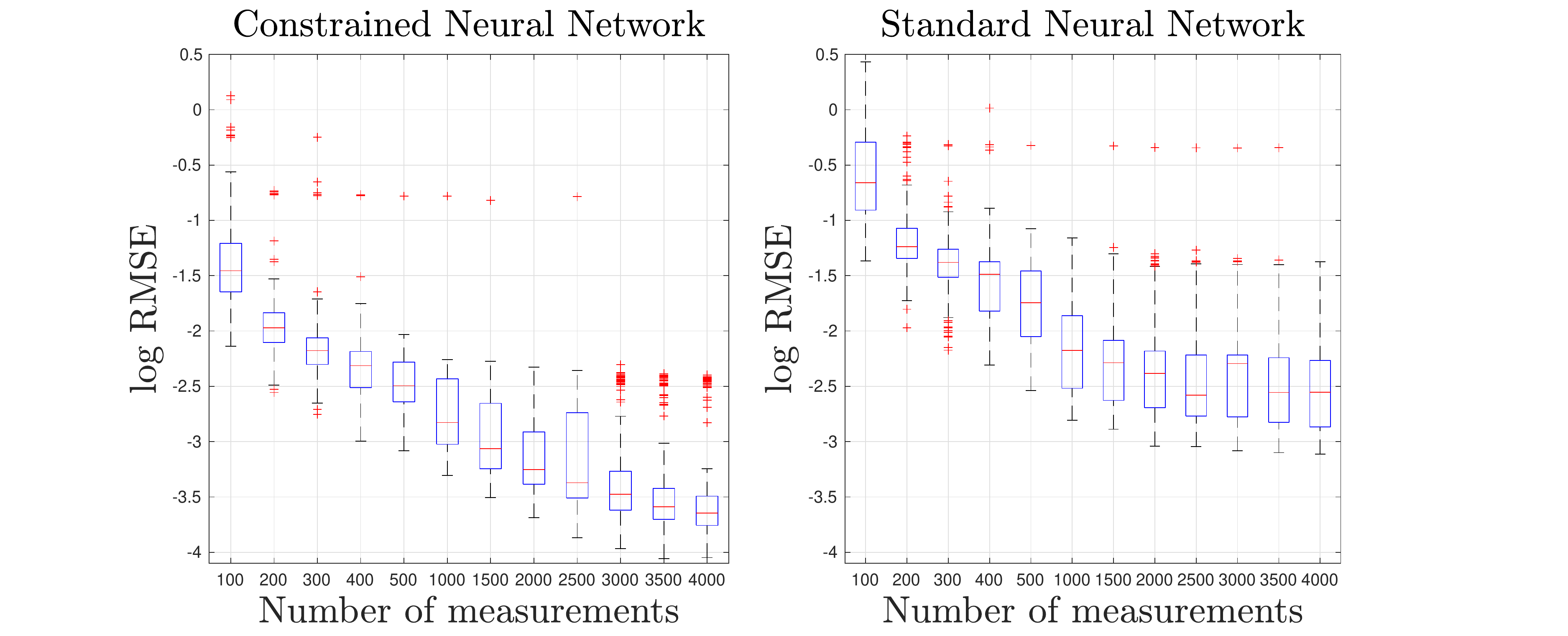}}
    \hspace{5mm}
    \subcaptionbox{Network size study\label{fig:net_size_study}}{\includegraphics[trim=90 5 150 5, clip,width=0.45\linewidth]{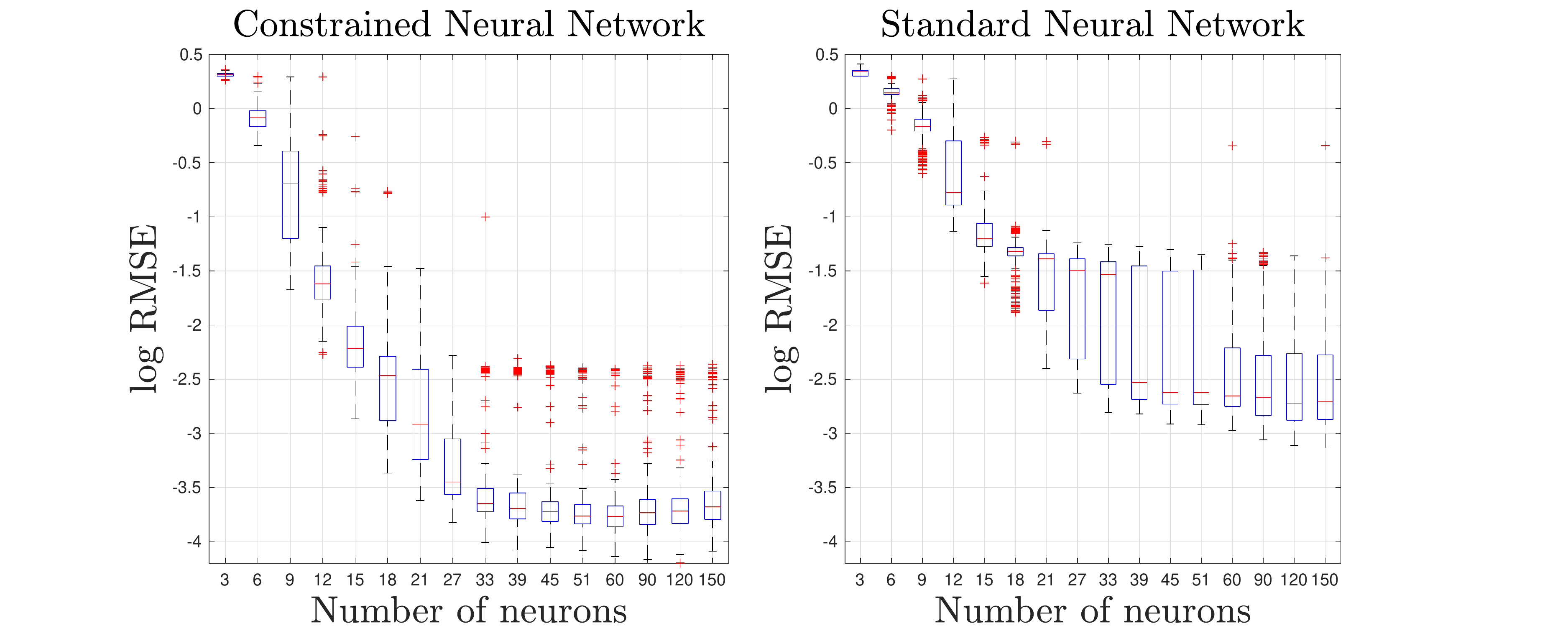}}
    \caption{Two studies comparing the performance of the proposed constrained neural network based model with a standard unconstrained neural network using simulated measurements of a divergence-free field. The RMSE is compared as (\subref{fig:n_data_study}) the number of measurements is increased and (\subref{fig:net_size_study}) as the size of the neural network is increased. }
    \label{fig:convergence_studies}
\end{figure*}

The regression of this problem using the proposed constrained neural network and an unconstrained (standard) neural network is compared. The networks root mean square error (RMSE) are compared in two studies:
\begin{enumerate}
    \item Maintaining a constant network size of 2 hidden layers (100 neurons in the first and 50 in the second) and increasing the number of measurements. See Figure~\ref{fig:n_data_study}.
    \item Maintaining a constant number of measurements (4000) and increasing the network size. In this case, the total number of neurons is reported with two-thirds belonging to the first hidden layer and one third belonging to the second. See Figure~\ref{fig:net_size_study}.
\end{enumerate}

In both studies, 200 random trials were completed with the measurements randomly picked over the domain $[0,4]\times[0,4]$, and corrupted by zero-mean Gaussian noise of standard deviation $\sigma=0.1$. For both networks, a \texttt{tanh} activation layer was placed on the output of the hidden layers. The networks were then trained using a mean squared error loss function and the ADAM optimiser, with the learning rate reduced as the validation loss plateaued.
A uniform grid of $20\times20$ points was chosen to predict the function values at. The root mean square error was then calculated between the true vector field and the predictions at these locations.
To focus this analysis on the impacts of the suggested approach, regularisation and other methods to reduce overfitting were not implemented. The effect of regularisation on both networks is considered in \ref{sub:regularisation_study}.

In both these studies, the proposed approach yields a significantly lower RMSE than a standard neural network. To highlight a few points, the proposed approach with 500 measurements has the same RMSE as the standard neural network with 4000 measurements. Similarly, with 21 total neurons the proposed approach performs as well as the standard neural network with 150 neurons.

\begin{figure*}[!ht]
    \centering
\includegraphics[trim=75 0 75 0, clip,width=0.75\linewidth]{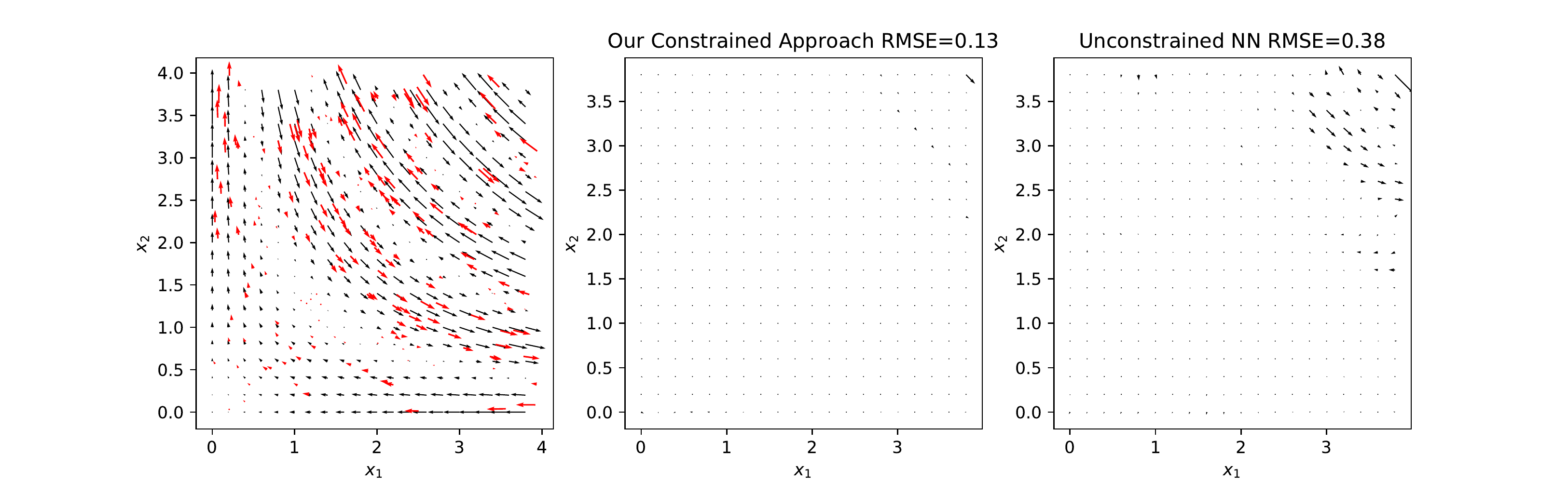}
    \caption{Comparison of learning the divergence-free field from 200 noisy observations using an unconstrained neural network (NN) and our constrained approach. Left: the true field (grey) and observations (red). Centre and right: learned fields subtracted from the true field. The comparison was performed using 2 hidden layers, 100 neurons in first, 50 in second for both methods.}
    \label{fig:divergence_free_fields}
\end{figure*}

An example of the learned vector fields from 200 noisy observations is provided in Figure~\ref{fig:divergence_free_fields}. For this comparison, both networks had two hidden layers with 100 neurons in the first and 50 in the second and a \texttt{tanh} activation function was placed on the outputs of both hidden layers. 

\begin{figure*}[!ht]
    \centering
\includegraphics[trim=0 0 0 0, clip,width=0.8\linewidth]{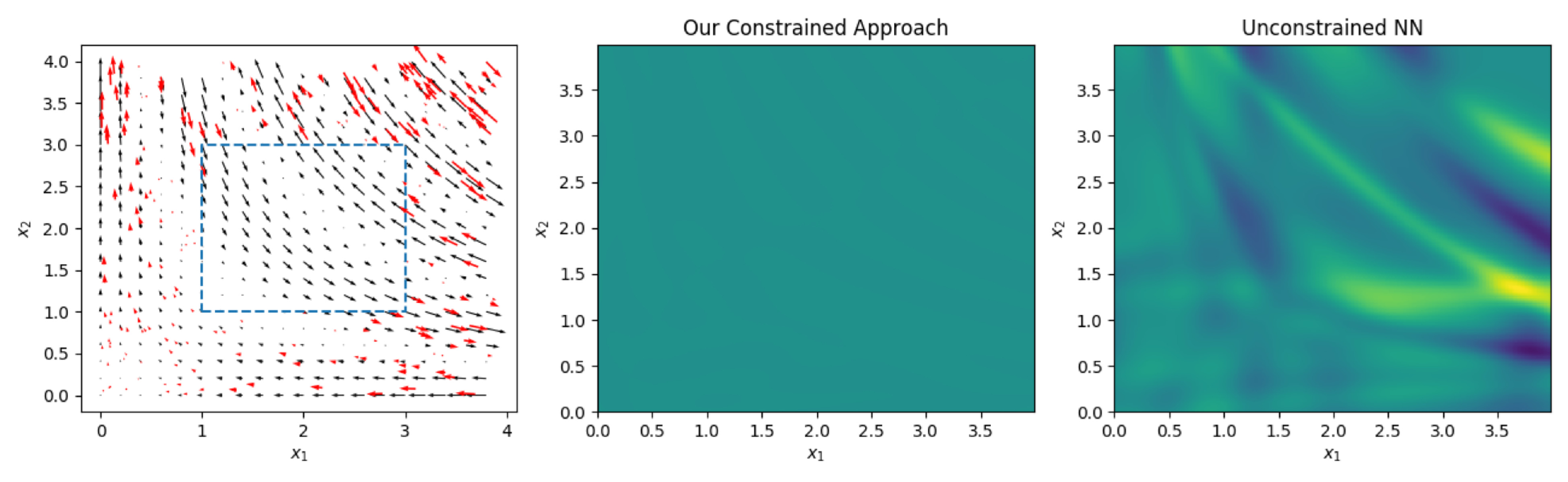}
    \caption{Comparison of learned fields constraint violations from 200 simulated noisy observations of a divergence-free field using an unconstrained neural network and our approach. Left: the true field (grey) and observations (red). Centre and right: the constraint violations. No measurements were made inside the dashed blue box. Centre and right: Constraint violations for the learned fields calculated as $c = \pd{f_1}{x}+\pd{f_2}{x_2}$.}
    \label{fig:constraint_violations}
\end{figure*}

These results indicate that the proposed approach can achieve equivalent performance with either less data or smaller network size.
Another property of our constrained neural network is that its predictions will automatically satisfy the constraints. This is true even in regions where no measurements have been made as illustrated in Figure~\ref{fig:constraint_violations}. By comparison, the standard neural network gives estimates which violate the constraints.

\subsection{Simulated Divergence-Free Function Continued} 
\label{sub:simulated_divergence_free_function_continued}
In the previous section, our proposed approach was compared against a neural network that made no attempt to incorporate knowledge of the constraints. In this section, we compare the performance of a neural network that augments the loss function by penalising the constraint violation at a finite number of points. Similar versions of this relatively straightforward approach have been used to approximate solutions to differential equations \citep{dissanayake1994neural,sirignano2018dgm,raissi2017physics,raissi2017physics2} and to learn plasma fields subject to equilibrium constraints \citep{van1995neural}. 

The simulated divergence-free function, equation \eqref{sub:simulated_divergence_free_function}, is learned using a neural network with an augmented loss function given by
\begin{equation}
  \text{loss} = \frac{1}{N}\sum_{i=1}^N (y_i-\hat{y}_i)^2 + \lambda\frac{1}{N_c}\sum_{j=1}^{N_c}|c_j|, 
\end{equation}
where $\hat{y}_i$ is the neural network prediction of measurement $y_i$, $N_c$ is the number of points the constraint violation is evaluated at, the constraint violation is given by 
\begin{equation}
  c = \pd{f_1}{x}+\pd{f_2}{x_2},
\end{equation}
and $\lambda$ is a tuning parameter that weights the relative importance of the measurements and the constraint. Note that varying the ratio of measurement points to constraint evaluation points has a similar effect to changing $\lambda$. 

\begin{figure}[ht]
    \centering
      \subcaptionbox{\label{fig:pc_rmse}}{\includegraphics[trim=5 5 50 5, clip,width=0.325\linewidth]{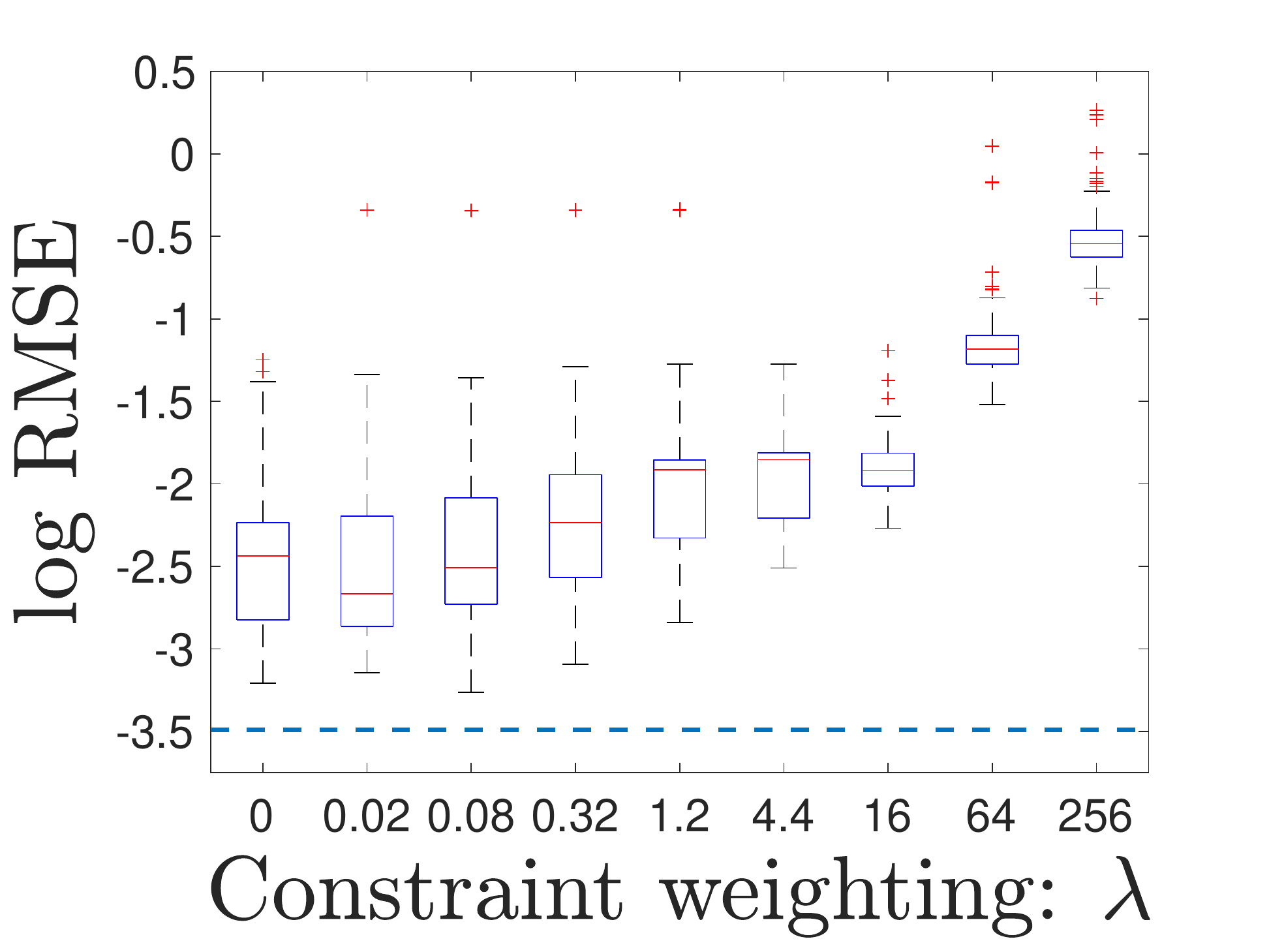}}
    \hspace{5mm}
    \subcaptionbox{\label{fig:pc_mac}}{\includegraphics[trim=5 5 50 5, clip,width=0.325\linewidth]{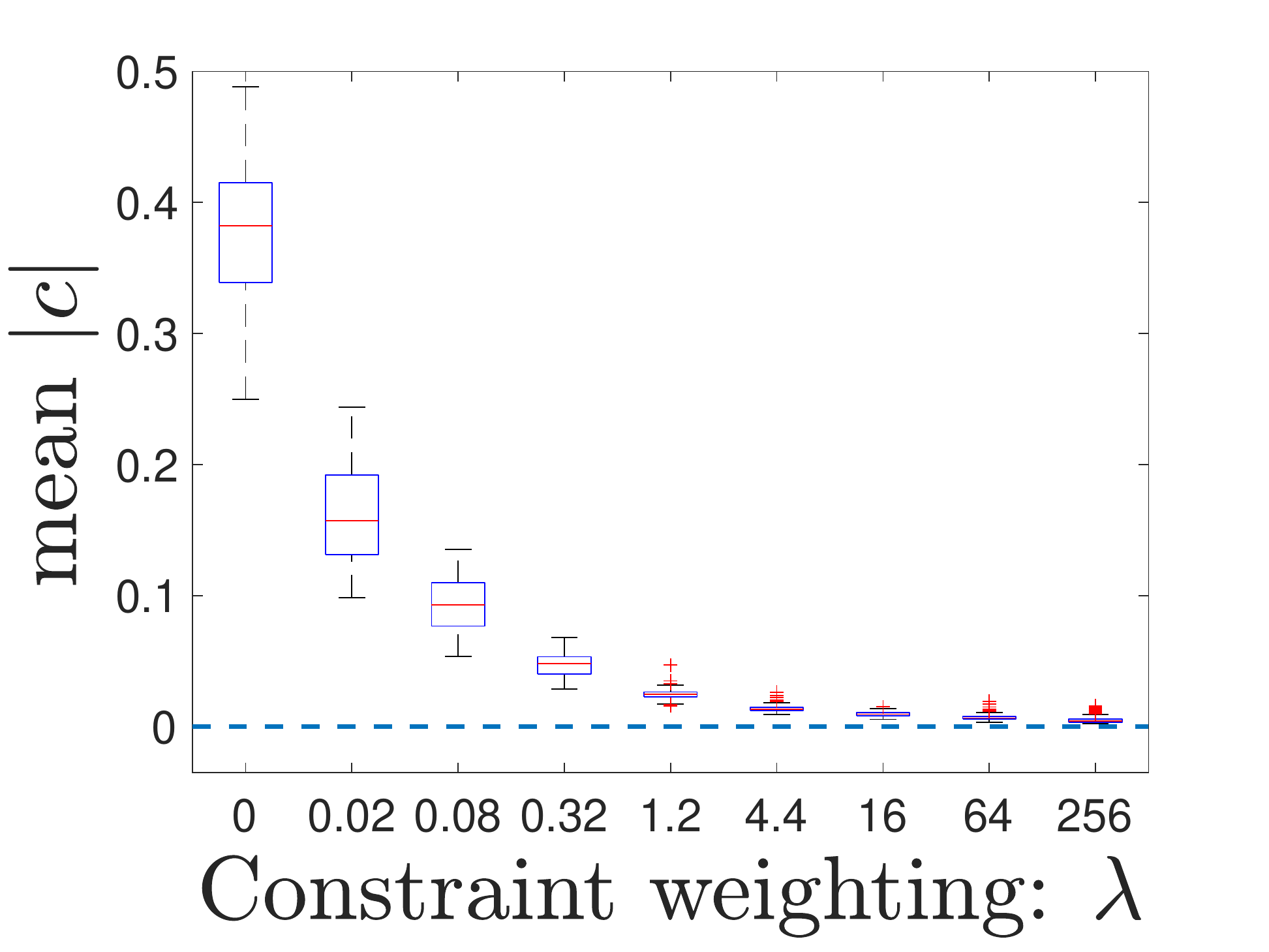}}
    \caption{Performance of a neural network with cost function augmented by penalising the mean squared constraint violation. The study compares (\subref{fig:pc_rmse}) the RMSE of the predicted field and (\subref{fig:pc_mac}) the mean absolute constraint violation of the predicted field for a range of weighting factors $\lambda$. For comparison, results from our proposed approach for the same number of measurements is indicated by the dashed line.}
    \label{fig:point_constraint_study}
\end{figure}

The results from $200$ random trials with $N=3000$, $N_c=3000$ and $\lambda$ ranging from $0$ to $256$ are shown in Figure~\ref{fig:point_constraint_study}. Also indicated, by a dashed line, is the median result using our proposed constrained approach for the same number of measurements.
These results indicate that rather than improving the neural network's predictions this approach of augmenting the cost function creates a trade-off between learning the field from the measurements and achieving low constraint violation.
In contrast, our proposed approach removes the need to tune the weight $\lambda$ by building the constraints into the model. Building the constraints into the model also has the added benefit of reducing the problem size, which improves the predictions and guarantees the constraints to be satisfied at all locations.

\subsection{Simulated Strain Field} 
\label{sub:simulated_strain_field}
An example of a more complex constraint is given by considering the estimation of strain fields.
Strain fields describe the relative deformation of points within a solid body and can be measured by neutron and X-ray diffraction \citep{noyan87} providing a means to study the stress---a quantity that cannot be directly measured.
Maximum stresses are commonly accepted as a major contributing factor to component failure \citep {sadd2009elasticity} and hence studying stress is of interest for the design of engineering components.

Physical strain fields satisfy the equilibrium constraints and, as such, it is important to ensure that any estimates of these fields from measurements also satisfies these constraints \citep{sadd2009elasticity}. 
Here, we consider a two-dimensional strain field with components described by $\epsilon_{xx}(x,y),\, \epsilon_{yy}(x,y),\, \epsilon_{xy}(x,y)$. 
Under the assumption of plane stress the equilibrium constraints are given by \citep{gregg2018tomographic}
\begin{equation}
  \begin{split}
    \pd{}{x}(\epsilon_{xx}+\nu\epsilon_{yy}) + \pd{}{y}(1-\nu)\epsilon_{xy} &= 0, \\
    \pd{}{y}(\epsilon_{yy}+\nu\epsilon_{xx}) + \pd{}{x}(1-\nu)\epsilon_{xy} &= 0. 
  \end{split}
\end{equation}
A neural network based model satisfying these constraints can be derived from physics using the so-called Airy stress function \citep{sadd2009elasticity} and is given by
\begin{equation}
  \pmat{{\epsilon}_{xx}\\ {\epsilon}_{yy}\\{\epsilon}_{xy}} = \pmat{\pdd{}{y} - \nu\pdd{}{x}\\
                          \pdd{}{x} - \nu\pdd{}{y}\\
                          -(1+\nu)\pdt{}{x}{y}}g.
\end{equation}

This model is used to learn the classical Saint-Venant cantilever beam strain field under an assumption of plane stress \citep{beer2010mechanics} from 200 noisy simulated measurements. Details of this strain field and the measurements is given in \ref{sec:strain_field_and_measurement_details}.

\begin{figure*}[!ht]
    \centering
\includegraphics[trim=0 0 0 0, clip,width=0.6\linewidth]{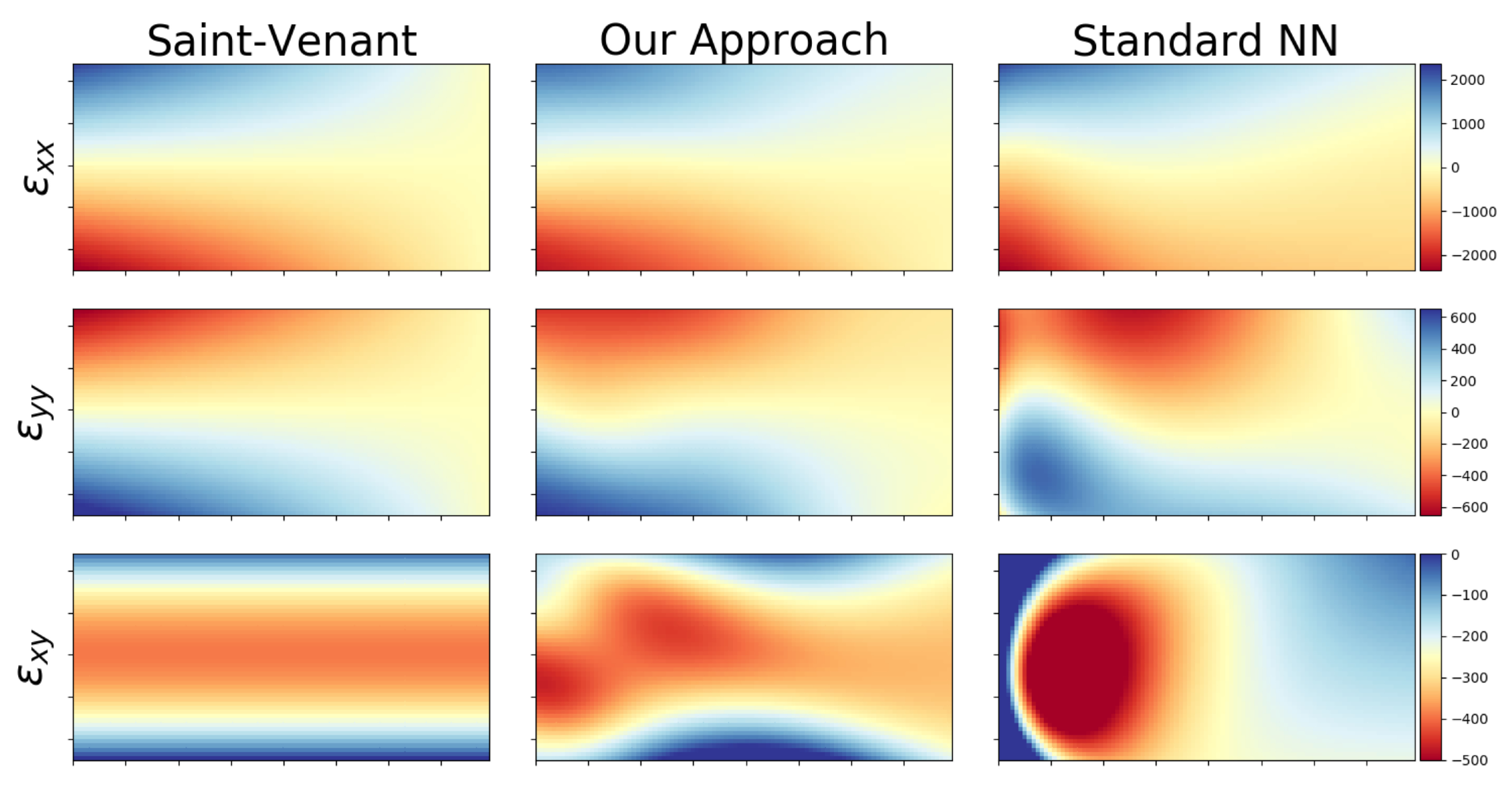}
    \caption{The theoretical Saint-Venant cantilever beam field and strain fields learned from 200 noisy measurements using the presented constrained approach and a standard neural network. Both networks have three hidden layers with 20, 10, and 5 neurons, respectively. Values are given in micro strain.}
    \label{fig:strain_field}
\end{figure*}

Predictions of this strain field using the proposed model and a standard neural network are shown in Figure~\ref{fig:strain_field}. 
The proposed approach gives an RMSE of $\num{5.52e-5}$ compared to $\num{67.7e-5}$ for the standard neural network.
Qualitatively, we can see that the proposed approach provides more accurate estimates of the strain field, particularly for the $\epsilon_{xy}$ component.

\subsection{Real Data} 
\label{sub:real_data}

Magnetic fields can be mathematically described as a vector field mapping a 3D position to a 3D magnetic field vector, $\mathbf{B}$. Based on the magnetostatic equations, this can be modelled as a curl-free vector field \citep{wahlstrom2015modeling,solin2014hilbert}:
\begin{equation}
    \nabla \times \mathbf{B} = \mathbf{0}.
\end{equation}
As such, a neural network satisfying the curl-free constraint can be designed to model the magnetic field according to
\begin{equation}
    \widehat{\mathbf{B}} = \pmat{\pd{}{x_1} \\ \pd{}{x_2} \\ \pd{}{x_3}}g.
\end{equation}
With a magnetic field sensor and an optical positioning system, both position and magnetic field data have been collected in a magnetically distorted indoor environment --- with a total of 16,782 data points collected.
Details of the data acquisition can be found in the supplementary materials of Jidling et. al.~\citep{jidling2017linearly}, where this data was previously published. 
Figure~\ref{fig:mag_predictions} illustrates magnetic field predictions using a constrained neural network trained on 500 measurements sampled from the trajectory shown in black.
The constrained neural network had two hidden layers of 150 and 75 neurons, with \texttt{Tanh} activation layers.
Using the remaining data points for validation, our constrained model has a RMSE validation loss of $0.048$ compared to $0.053$ for a standard unconstrained neural network of the same size and structure.

\begin{figure*}[ht]
    \centering
    \subcaptionbox{Data size study\label{fig:mag_n_data}}{\includegraphics[trim=90 5 150 5, clip,width=0.45\linewidth]{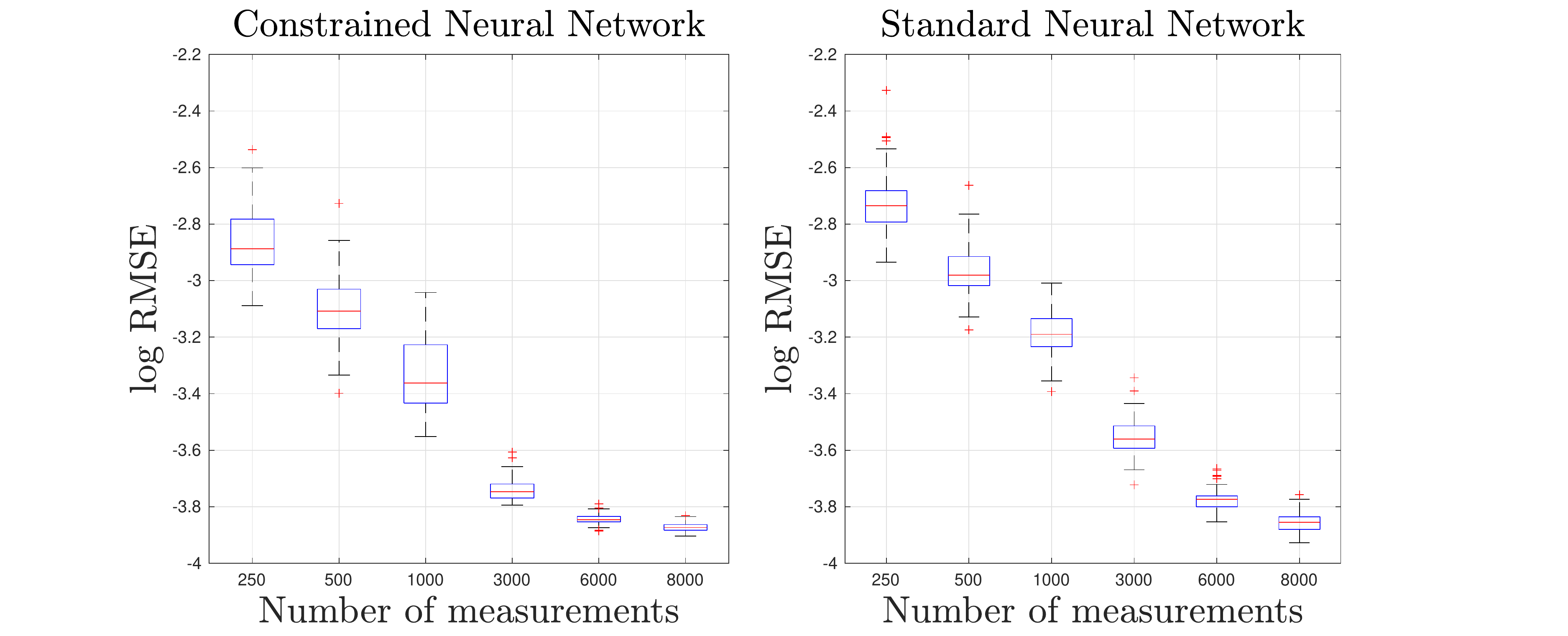}}
    \hspace{5mm}
    \subcaptionbox{Network size study\label{fig:mag_net_size}}{\includegraphics[trim=90 5 150 5, clip,width=0.45\linewidth]{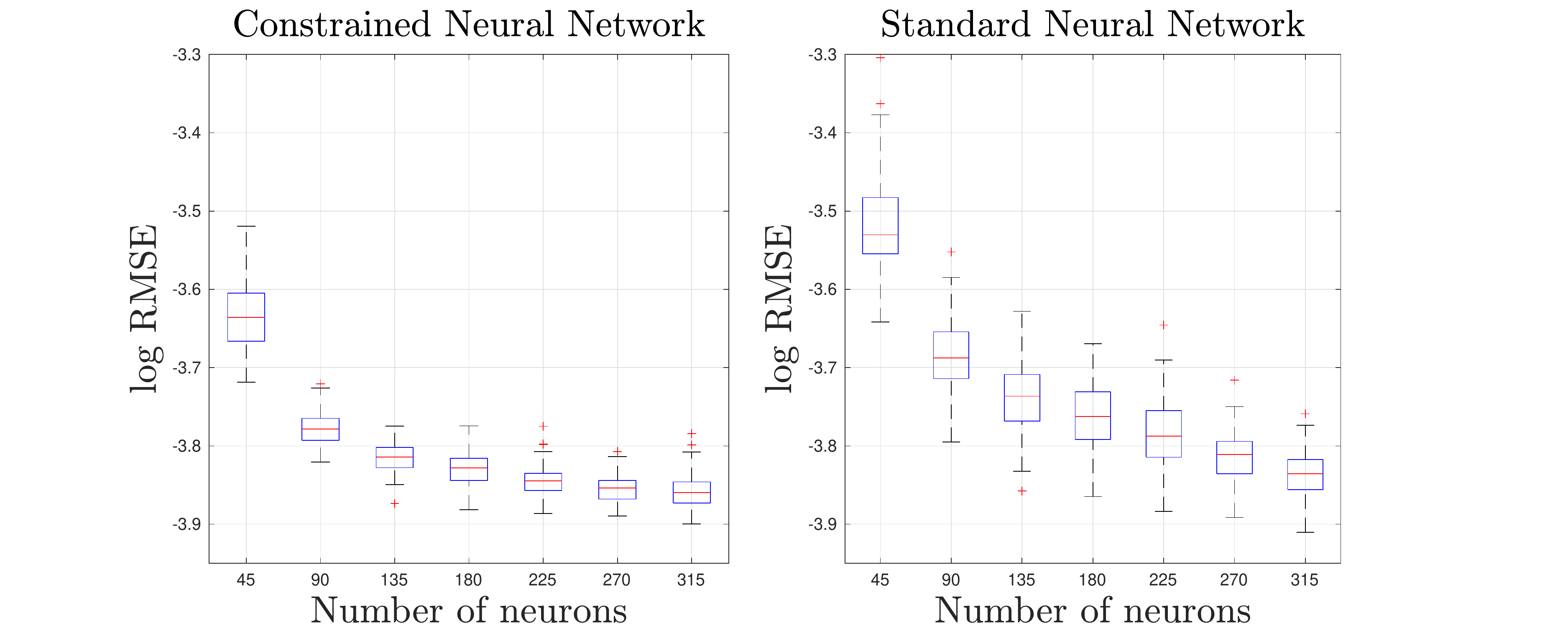}}
    \caption{Two studies comparing the performance of the proposed constrained neural network based model with a standard unconstrained neural network using data collected of a magnetic field. The RMSE is compared as (\subref{fig:mag_n_data}) the number of measurements is increased and (\subref{fig:mag_net_size}) as the size of the neural network is increased.}
    \label{fig:mag_convergence_studies}
\end{figure*}

Two studies were run comparing the proposed approach and a standard neural network for a range of training data sizes and neural network sizes. 
The networks RMSE when validated against 8,000 reserved validation data points are shown in Figure~\ref{fig:mag_convergence_studies} for the following settings:
\begin{enumerate}
    \item Maintaining a constant network size of 2 hidden layers (150 neurons in the first and 75 in the second) and increasing the number of measurements. See Figure~\ref{fig:mag_n_data}.
    \item Maintaining a constant number of measurements (6000) and increasing the network size. In this case, the total number of neurons is reported with two-thirds belonging to the first hidden layer and one third belonging to the second. See Figure~\ref{fig:mag_net_size}.
\end{enumerate}
For both studies, the results of training the networks for 100 random initialisations are shown.

The studies show that the proposed approach performs better than the standard neural network for a smaller number of measurements or a smaller network size. 
As the number of measurements or neurons is increased, the performance of both networks converges. 
This is expected as given enough measurements and a large enough network size, both methods should converge to the true field and hence a minimum validation RMSE.

\section{Related work} 
\label{sec:related_work}
Focusing on neural networks, related work falls into two broad categories: incorporating known physics relations as prior knowledge, and optimising neural networks subject to constraints. Here, we discuss some of these methods that are closely related or of particular interest.

Several papers have discussed incorporating differential equation constraints into neural network models by augmenting the cost function to include a penalty term given by evaluating the constraint at a finite number of points \citep{van1995neural,dissanayake1994neural,sirignano2018dgm,raissi2017physics}. 
This idea is presented as a method to approximate the solution to partial differential equations by Dissanayake and Phan-Thien~\citep{dissanayake1994neural}, as it transforms the problem into an unconstrained optimisation problem.
Van Milligen et al.~\citep{van1995neural} applies this idea to learning plasma fields which are subject to equilibrium. 
Sirignano and Spiliopoulos~\citep{sirignano2018dgm} applies this idea to the learning of high dimensional partial differential equations, with boundary conditions included in the same manner as the constraints --- that is, by augmenting the cost function.

A similar approach was used by Raissi et al.~\citep{raissi2019physics} to learn the solution to linear and non-linear partial differential equations and they demonstrate the approach on examples from physics such as Shr\"odinger's equation.
They also demonstrated that automatic differentiation included in TensorFlow and PyTorch provides a straightforward way to calculate the augmented cost.
Further, they provide an alternate method for discovering ordinary differential equation style models from data.

\red{Outside of modelling vector fields and differential equations, augmenting the cost function to penalise constraint violation has seen use in other applications of neural networks including image segmentation and classification \cite{pathak2015constrained,jia2017constrained,liu2018constrained,oktay2017anatomically}. 
An alternative that strictly enforces the constraints is to learn the neural network  parameters  by  solving  a  soft-barrier  constrained  optimisation  problem \cite{marquez2017imposing,kervadec2019log}. However, it has been shown that the soft-barrier style approaches are outperformed by the augmented cost function approaches \cite{drgona2020physics}, possibly due to the challenges of solving the non-convex optimisation problem \cite{yang2019advancing}.}

While the approach of augmenting the cost function can be used to solve the problem presented in our paper it has some downsides. Firstly, it does not guarantee that the constraint is satisfied, and this is especially true in regions where the constraint may not have been evaluated as part of the cost. 
Secondly, its performance is subject to the number of points at which the constraint is evaluated.
Thirdly, a relative weighting needs to be chosen between the original cost and the cost due to constraint satisfaction and this creates a trade-off.
The approach presented in our paper avoids these issues by presenting a neural network model that uses a transformation to guarantee the constraints to be satisfied everywhere rather than augmenting the cost function. 
\red{In Section~\ref{sub:simulated_divergence_free_function_continued}, we have compared to performance of our proposed approach to that of augmenting the  cost function.}

Augmenting the cost function does, however, have the advantage that it can be used even when no suitable transform $\mathcal{G}_\mathbf{x}$ is forthcoming. For instance, it can be used to enforce boundary constraints \citep{sirignano2018dgm,raissi2017physics}. Since using our proposed approach does not exclude the possibility of also augmenting the loss function, we suggest that the two approaches are complementary. Whereby, our proposed approach is used to satisfy constraints for which a suitable transform $\mathcal{G}_\mathbf{x}$ can be designed, and then the cost function is augmented to penalise violation of other constraints, such as boundary conditions.
A similar combined approach was used to model strain fields subject to both equilibrium constraints and boundary conditions \citep{hendriks2018traction}.

The idea of modelling potential functions using neural networks as a means to include prior knowledge about the problem is not new and has been used to learn models of dynamic systems and vector fields.

There are several examples of using neural networks to learn the Hamiltonian or Lagrangian of a dynamic system and training this model using the derivatives of the neural network \citep{greydanus2019hamiltonian,Lutter2019DeepLN,zhong2019symplectic,gupta2019general,cranmer2020lagrangian,lutter2019deep}.
These methods ensure that the learned dynamics are conservative, i.e. the total energy in the system is constant.
Another approach to learning dynamic systems is presented by Chen et al.~\citep{chen2018neural} where neural networks are used to model solutions to ODEs and Massaroli et al.~\citep{massaroli2019port} propose a Port-Hamiltonian based approach to training these models.

A method for simultaneous fitting of magnetic potential fields and force fields using neural networks is studied by Pukrittayakamee et al.~\citep{pukrittayakamee2009simultaneous}. 
This method uses a neural network to model the potential field and the force field is then the partial derivatives of the neural network.
In their work, measurements of both the potential field and the force field are used to train the model.
This work was later extended to provide a practical approach to fitting measurements of a function and its derivatives using neural networks as well as a discussion on specific types of overfitting that might be encountered \citep{pukrittayakamee2011practical}.
A similar approach is taken by Handley and Popelier~\citep{handley2010potential} where it is used for modelling molecular dynamics and Monte Carlo studies on gas-phase chemical reactions.

Although it was not the motivation for the method, the model for the force field given by Pukrittayakamee et al.~\citep{pukrittayakamee2009simultaneous} will obey the curl-free constraint and is equivalent to that presented in Section~\ref{sub:real_data}.
In our work, we extend the idea of representing the target function by a transformation of a potential function modelled by a neural network to a broader range of problems that obey a variety of constraints. 
Additionally, we focus only on the transformed target function and do not require measurements of the potential function.


Another interesting idea is presented by Schmidt and Lipson~\citep{schmidt2009distilling} who derived a method for distilling natural laws from data.
In their work, symbolic terms including partial derivatives are used as building blocks with which to learn equations that the data satisfies.

Instead of incorporating known physics or constraints into neural networks, another area of research is using neural networks to solve constrained optimisation problems. 
Several methods using neural networks to solve constraint satisfaction problems have been presented \citep{adorf1990discrete,tsang1992generic}. 
For example, Xia et al.~\citep{xia2002projection} used a recurrent neural network for solving the non-linear projection formulation, and is applicable to many constrained linear and non-linear optimisation problems.
Similar to our approach, their method uses a projection or transformation of the neural network; however, both the motivation and realisation are substantially different.
Neural networks have also been applied to solving optimisation problems quadratic cost functions subject to bound constraints \cite{bouzerdoum1993neural}.

\section{Conclusion} 
\label{sec:conclusion}
An approach for designing neural network based models for regression of vector-valued signals in which the target function is known to obey linear operator constraints has been proposed.
By construction, this approach guarantees that any prediction made by the model will obey the constraints for any point in the input space.
It has been demonstrated on simulated data and real data that this approach provides benefits by reducing the size of the problem and hence providing performance benefits.
This reduces the required number of data points and the size of the neural network --- providing savings in terms of time and cost required to collect the data set.

\red{It was also demonstrated that the proposed approach can out-perform the commonly used method of augmenting the cost function to penalise constraint violations.
This is likely due to the fact that augmenting the cost function increases the size of the problem (by including artificial measurements of the constraint) and creates a trade-off between constraint satisfaction and minimising error. Whereas our proposed approach reduces the size of the problem and does not suffer from this trade-off.}

The proposed approach constructs the model by a transformation of an underlying potential function, where the construction is chosen such that the constraints are always satisfied. 
This transformation may be known from physics or, in the absence of such knowledge, constructed using a method of ansatz.
Additionally, we provide an example of extending this approach to affine constraints (constraints with a non-zero right-hand side) in \ref{sec:simulated_affine_example}.
\red{Whilst the proposed approach admits a wide class of neural networks to model the underlying function, the work presented in this paper focusses on FCNs and it would be interesting future work to explore the use of other types of neural networks such as CNNs.}

Another interesting area for future research would be to determine if it is possible to learn the transformation as a combination of symbolic elements using tools similar to those presented by Schmidt and Lipson~\citep{schmidt2009distilling}.

\section*{Acknowledgements} 
\label{sec:acknowledgements}
This research was financially supported by the Swedish Foundation for Strategic Research (SSF) via the project \emph{ASSEMBLE} (contract number: RIT15-0012) and by the Swedish Research Council via the project \emph{Learning flexible models for nonlinear dynamics} (contract number: 2017-03807).

\appendix

\section{Regularisation Study} 
\label{sub:regularisation_study}
In the simulated study in Section~\ref{sub:simulated_divergence_free_function}, the use of regularisation was not considered. Here, we consider the impact of regularisation on the performance of both the standard and the constrained neural network. 
To investigate this impact the network size study from the previous section was rerun with the networks regularised by the addition of an L2 penalty on the network weights to the loss function,
\begin{equation}
  \text{loss} = \frac{1}{N}\sum_i^N(y_i-\hat{y}_i)^2 + \gamma \sum_j^m w_j^2,
\end{equation}
where $\hat{y}_i$ is the network's prediction of the measurement $i$, $w_j, \ \ \forall j=1,\dots,m$ are the network weights, and $\gamma$ is a tuning parameter often known as the weight decay.

\begin{figure}[tb]
  \centering
  \includegraphics[width=0.5\linewidth]{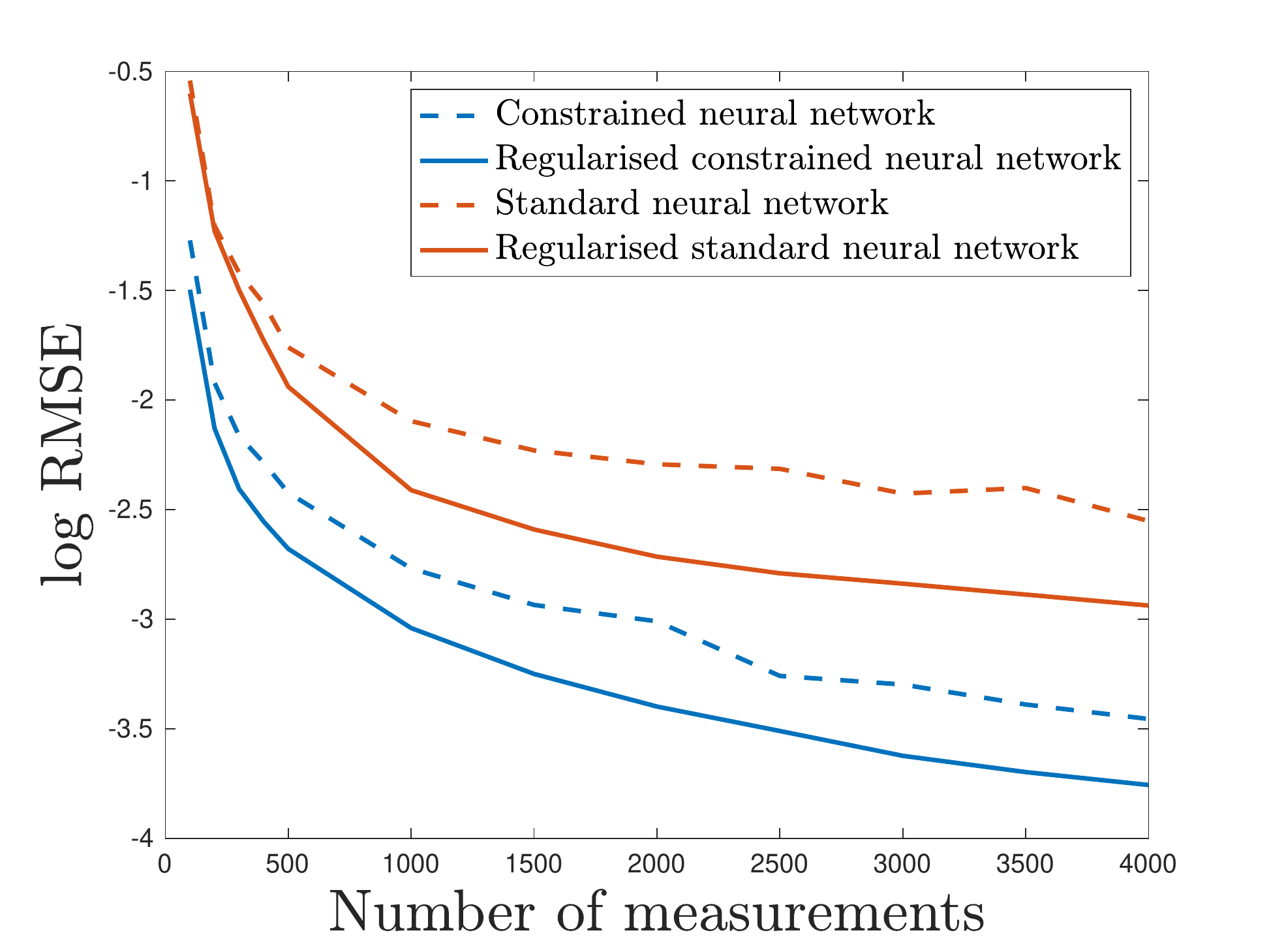}
  \caption{Comparison of both the constrained and standard neural network with and without regularisation. A weight decay of $\gamma=1\times10^{-4}$ is used to regularise both networks. The average loss for 200 random trials is shown.}
  \label{fig:regularisation}
\end{figure}

The impact of regularisation on the performance of both networks is shown in Figure~\ref{fig:regularisation}. A weight decay of $\gamma=1\times10^{-4}$ was used to regularise both networks as it was found to give the best results for the standard neural network.
These results indicate that the inclusion of regularisation can improve the performance of both the standard and the proposed constrained neural network, in this case, by approximately the same amount.
This comparison further highlights the benefit of the constrained approach as even without regularisation it performs better than the regularised standard neural network.

\section{Simulated Affine Example} 
\label{sec:simulated_affine_example}
It is also possible to design a model to satisfy an affine constraint $\mathcal{C}_\mathbf{x}\mathbf{f} = \mathbf{b}$. 
This type of constraint arises for vector fields that have a prescribed divergence of curl. For example, Maxwell's equations \citep{fleisch2008student} and low-mach number flow \citep{almgren2006low}.

To illustrate the approach, we design a neural network that will satisfy constant divergence, i.e. $\nabla \mathbf{f} = b$, and demonstrate the method in simulation.
The model satisfying this constraint can be built by starting from the model used in the previous section that satisfies $\nabla \mathbf{f} = 0$.
From this starting model, we need to add a component that when mapped through the constraints results in a constant term.
This is easily achieved using a 2 input 1 output linear layer with no bias term, giving the final model as 
\begin{equation}
    \hat{\mathbf{f}} = \pmat{\pd{}{x_2} \\ -\pd{}{x_1}}\mathbf{g} + c_0x_1 + c_1 x_2,
\end{equation}
where the weights $c_0$ and $c_1$ will be learned along with the rest of the neural network parameters.

\begin{figure*}[!ht]
    \centering
\includegraphics[trim=0 0 0 0, clip,width=0.9\linewidth]{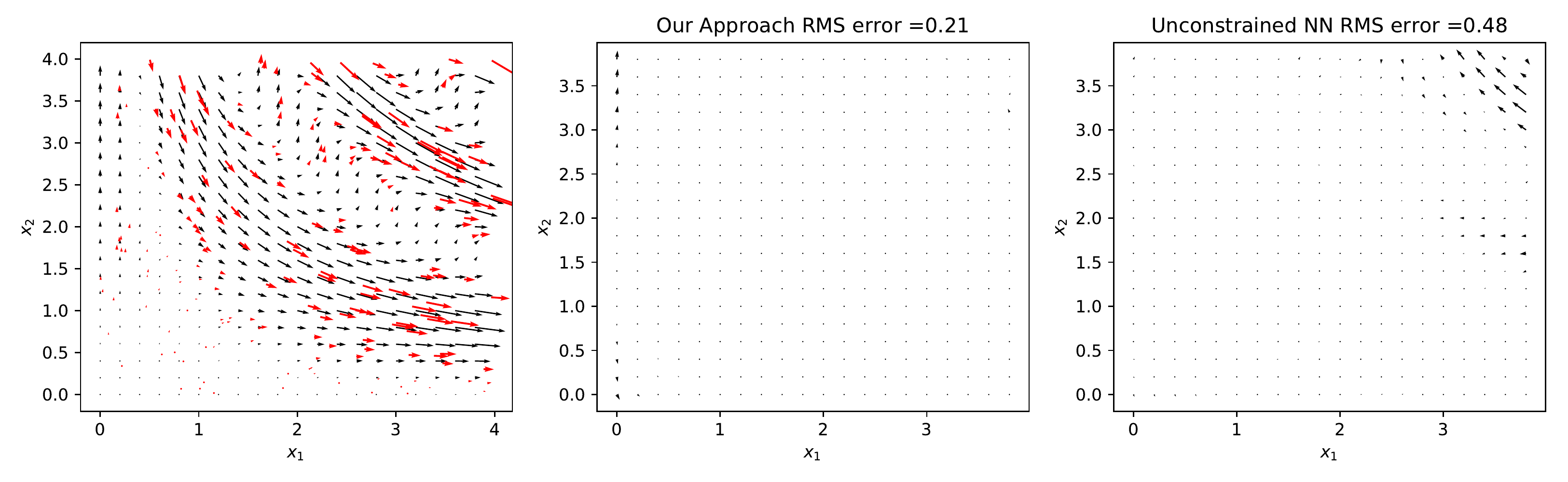}
    \caption{Comparison of learning the affine constrained field from 200 noisy observations using an unconstrained neural network and our approach. Left: the true field (grey) and observations (red). Centre and right: learned fields subtracted from the true field. Done using 2 hidden layers, 100 neurons in first, 50 in second for both methods.}
    \label{fig:affine_example}
\end{figure*}

Figure~\ref{fig:affine_example} shows the results of learning a field satisfying these constraints using our approach and a standard neural network.
Measurement locations are randomly picked over the domain $[0,4]\times[0,4]$,
with 200 measurements simulated from the field given by
\begin{equation}
\begin{split}
    f_1(x_1,x_2) &= \exp(-ax_1x_2)ax_1\sin(x_1x_2)\\&\hspace{10mm}-\exp(-ax_1x_2)x_1\cos(x_1x_2) + 1.1x_1, \\
    f_2(x_1,x_2) &= \exp(-ax_1x_2)x_2\cos(x_1x_2)\\&\hspace{10mm}-\exp(-ax_1x_2)ax_2\sin(x_1x_2) -0.3x_2,
\end{split}
\end{equation}

and zero-mean Gaussian noise of standard deviation $\sigma=0.1$ added. For both networks, two hidden layers are used with 100 and 50 neurons respectively and \texttt{Tanh} activation layers.
The field is then predicted at a grid of $20\times20$ points with the proposed approach achieving a RMSE of 0.21 compared to 0.48 for the standard neural network.

\section{Strain Field and Measurement Details} 
\label{sec:strain_field_and_measurement_details}
This section provides details of the strain field equations and the simulated measurements used for the simulated strain field example in the main paper.
The simulation uses the classical Saint-Venant cantilever beam strain field equations under an assumption of plane stress \citep{beer2010mechanics};
\begin{equation}
  \begin{split}
    \epsilon_{xx}(x,y) &= \frac{P}{EI}(l-x)y, \\
    \epsilon_{yy}(x,y) &= -\frac{\nu P}{EI}(l-x)y, \\
    \epsilon_{xy}(x,y) &= -\frac{(1+\nu)P}{2EI}\left(\left(\frac{h}{2}\right)^2-y^2\right),
  \end{split}
\end{equation}
where $P=\SI{2}{\kilo\newton}$ is the applied load, $E=\SI{200}{\giga\pascal}$ is the elastic modulus, $\nu=0.28$ is Poisson's ratio, $l=\SI{20}{\milli\meter}$ is the beam length, $h=\SI{10}{\milli\meter}$ is the beam height, $t=\SI{5}{\milli\meter}$ is the beam width, and $I=\frac{th^3}{12}$ is the second moment of inertia.

Simulated measurements of this strain field were made at random locations within the beam and were corrupted by zero-mean Gaussian noise with standard deviation of $\num{2.5e-4}$.
In practice, such measurements can be made by X-ray or neutron diffraction and correspond to the average strain within a small volume of material inside the sample, known as a gauge volume \citep{noyan87}. Gauge volumes can be made small enough that it is practical to treat these measurements as corresponding to points in the sample, and noise levels as low as $\num{1e-4}$ or better can be achieved.

\bibliography{references}

\end{document}